\documentclass{article}

    \PassOptionsToPackage{numbers, compress}{natbib}

 \usepackage[final]{neurips_bdl2019}

\usepackage[utf8]{inputenc} 
\usepackage[T1]{fontenc}    
\usepackage{hyperref}       
\usepackage{url}            
\usepackage{booktabs}       
\usepackage{amsfonts}       
\usepackage{nicefrac}       
\usepackage{microtype}      
\usepackage{times}

\usepackage{amsthm,amsmath,amssymb,amsfonts,graphicx,epsfig,pgf,subfig,rotating,longtable,booktabs, colortbl, graphics,fontenc}
\usepackage{algorithm}
\usepackage{algorithmic}
\usepackage{tabularx}
\usepackage{wrapfig}

\newcommand{\argmin}{\operatornamewithlimits{argmin}}
\newtheorem{thm}{Theorem}
\title{Refined $\alpha$-Divergence Variational Inference \\ via Rejection Sampling}

\author{
Rahul Sharma \\
IIT Kanpur \\
\texttt{rsharma@cse.iitk.ac.in}
\And
Abhishek Kumar \\
Google Research \\
\texttt{abhishk@google.com}
\And
Piyush Rai \\
IIT Kanpur \\
\texttt{piyush@cse.iitk.ac.in}
}

\begin{document}

\maketitle


\section{Introduction}
 We present an  approximate inference method, based on a synergistic combination of R\'enyi $\alpha$-divergence variational inference (RDVI) and rejection sampling (RS). RDVI is based on minimization of R\'enyi $\alpha$-divergence $D_\alpha(p||q)$ between the true distribution $p(x)$ and a variational approximation $q(x)$; RS draws samples from a distribution $p(x) = \tilde{p}(x)/Z_{p}$ using a proposal $q(x)$, s.t. $Mq(x) \geq \tilde{p}(x), \forall x$. Our inference method is based on a crucial observation that $D_\infty(p||q)$ equals $\log M(\theta)$ where $M(\theta)$ is the optimal value of the RS constant for a given proposal $q_\theta(x)$. This enables us to develop a \emph{two-stage} hybrid inference algorithm.
  
There is an increasing interest in developing more expressive variational posteriors for (shallow/deep) latent variable models and Bayesian neural networks~\cite{rezende2015variational,salimans2015markov,chen2016variational}. In particular, the combination of MCMC and variational methods have been used in recent work to learn expressive variational posteriors ~\cite{salimans2015markov} having the best of both worlds. Rejection Sampling~\cite{bishop2006pattern}, which we use as a subroutine (with learned $M$) in our algorithm $\alpha$-DRS, is a popular sampling technique that generates independent samples from a complex distribution indirectly through a simple distribution. In addition to being a useful sampling algorithm in its own right, recently \emph{approximations} of Rejection Sampling have also been used for designing variational inference algorithms. In particular, Variational Rejection Sampling (VRS) ~\cite{grover2018variational}, which uses rejection sampling to learn a better variational approximation. Recently Rejection sampling has also been used to improve the generated samples from GAN (Generative Adversarial Nets)~\cite{azadi2018discriminator} and improve priors for variational inference ~\cite{bauer2018resampled}.

\section{Connecting Rejection Sampling with R\'enyi $\alpha$-Divergence} 
\label{section_RS}
We now show how R\'enyi $\alpha$-divergence is related to rejection sampling, and how this connection can be leveraged to finetune the $q_\theta$ estimated by RDVI using $q_\theta$ as a proposal distribution of a rejection sampler, and generating a sample-based approximation of the exact distribution. The connection between R\'enyi $\alpha$-divergence and rejection sampling is made explicit by the following result

\begin{thm}
 When $\alpha$ $\to$ $\infty$, the R\'enyi $\alpha$ divergence becomes equal to the worst-case regret ~\cite[Theorem~6]{van2014renyi}.
\begin{equation}
   \lim_{\alpha \to \infty}  D_{\alpha}(p||q_\theta)= \text{log max$_{x\in \mathcal{X}}$}\frac{p(x)}{q_{\theta}(x)} \label{worst_case_regret}
\end{equation}
\end{thm}

It is interesting to note that $\lim_{\alpha \to \infty}  D_{\alpha}(p||q_\theta)$ in Eq. \eqref{worst_case_regret} is equal to the log of the optimal $M(\theta)$ value used in Rejection Sampling. It is easy to show that $q_{\theta}(x) \left(\text{max}_{x\in \mathcal{X}}\frac{p(x)}{q_{\theta}(x)}\right)\geq p(x)$, $\forall x \in \text{supp}(p(x))$.

In R\'enyi $\alpha$-divergence variational inference ~\cite{li2016renyi}, we learn the variational parameters $\theta$ such that the value of $\alpha$ divergence is minimized. Therefore, minimizing R\'enyi $\alpha$ divergence of $\infty$ order can serve the following purposes:
\begin{itemize}
    \item We can learn the optimal variational distribution $q_{\hat{\theta}}(x)$.
    \item We can learn the optimal value M($\hat{\theta}$) (expected number of iterations needed to generate one sample) such that rejection sampling could be performed with fewer rejections.
    \item The above rejection sampler can be used to ``refine'' $q_\theta$ using a sample-based approximation.
\end{itemize}

Although the above idea seems like an appealing prospect, optimizing R\'enyi $\alpha$ divergence of $\infty$ order is problematic. Instead of using Rejection Sampling for $\infty$ order $\alpha$-divergence, we will develop an approximate version of Rejection sampling for finite order $\alpha$-divergence.

\subsection{$\alpha$-Divergence Rejection Sampling}\label{approx_RS}
In this section, we summarize our algorithm $\alpha$-Divergence Rejection Sampling ($\alpha$-DRS) which augments the $\alpha$ divergence ~\cite{li2016renyi} method. The algorithm requires an input $\alpha$, the target distribution $p(x) = \tilde{p}(x)/Z_{p}$, and the variational distribution $q_{\theta}(x)$. Our algorithm $\alpha$-DRS consists of two stages.
\begin{itemize}
    \item In stage-1, given an input $\alpha$, we minimize the Monte-Carlo estimate of the exponentiated version of finite order $\alpha$-divergence ~\cite{dieng2017variational} with respect to the variational parameters $\theta$, i.e.,
    \begin{equation}
        \hat{\theta}= \argmin_{\theta} \frac{1}{S}\sum_{s=1}^{S}\left(\frac{\tilde{p}(x_{s})}{q_{\theta}(x_{s})}\right)^{\alpha}, \label{approx_alph_divg}
    \end{equation}
    Here $x_{s}$ are iid samples drawn from $q_{\theta}(x)$. 
    
    \item From stage-1, we learned the optimal $\hat{\theta}$. For the second stage we will learn $T$ from equation ~\eqref{quantile_funct} and perform approximate Rejection Sampling ~\eqref{accep_apprx} to learn a refined distribution $r_{\hat{\theta}}(x)$.
\end{itemize}

The acceptance probability for approximate RS is as follows:
\begin{equation}
a_{\hat{\theta}}(x|T)= 1/\left[1 +\left(\frac{q_{\hat{\theta}}(x)e^{-T}}{\tilde{p}(x)}\right) \right], \label{accep_apprx}
\end{equation}

where T is a hyperparameter controlling the acceptance rate.
\begin{thm}
For a fixed $\theta$, the approximate Rejection sampling always improves the R\'enyi $\alpha$ divergence between the estimated and actual posterior. The acceptance probability is approximated by equation ~\eqref{accep_apprx}. The proof of the theorem can be found in the supplementary material. 
\begin{equation}
    D_{\alpha}(p||r)\leq D_{\alpha} (p||q)  \label{final_deriv1} 
\end{equation}


\end{thm}

\subsection{Choosing the hyperparameter T}\label{approx_RS}
Although $D_{\alpha}(p||q)$ is a lower bound on $\log M(\hat{\theta})$ (property of $\alpha$-divergence), for high dimensions even this may be too large. The hyperparameter $T$ should be defined such that we can control the acceptance rate. Let's define $\mathcal{L}_{\theta}(x)=-\log \tilde{p}(x)+\log q_{\hat{\theta}}(x)$ where $x \sim q_{\hat{\theta}}(x)$, and redefine $T$ as
\begin{equation}
T=\left\{{\begin{matrix} - D_{\alpha}(p||q) &{\text{For low dimensions}} \\[8pt]
\mathcal{Q}_{\mathcal{L}_{\theta}(x)}(\gamma)&{\text{For high dimensions}}\end{matrix}}\right. \label{quantile_funct}
\end{equation},
where $\mathcal{Q}$ is quantile function defined over the random variable $\mathcal{L}_{\theta}(x)$ with hyperparameter $\gamma \in [0,1]$. The quantile function $\mathcal{Q}$ approach ~\cite{grover2018variational} allows us to select samples that have high-density ratios (similar to Rejection sampling) along with a well-defined acceptance rate (around $\gamma$ for most samples). Note that a similar methodology has been recently employed in Variational Rejection Sampling (VRS) \citep{grover2018variational} as well.




\section{Experiments}
In this section, we evaluate our proposed $\alpha$-DRS algorithm on synthetic as well as real-world datasets. In particular, we are interested in assessing the performance of $\alpha$-DRS as a method that can improve the variational approximation learned by RDVI.

\subsection{Gaussian Mixture Model Toy Example}

In this experiment,  we have chosen $p(x)$ to be a mixture of four Gaussian distributions.

\begin{center}
$p(x)=\frac{1}{4}\mathcal{N}(-12,0.64)+\frac{1}{4}\mathcal{N}(-6,0.64)+\frac{1}{4}\mathcal{N}(0,0.64)+\frac{1}{4}\mathcal{N}(6,0.64)$
\end{center}

\begin{wrapfigure}{c}{.4\textwidth}
\begin{minipage}{0.40\linewidth}
\centering
\setlength{\tabcolsep}{2pt}
  \begin{tabular}{c|c c c  c }
 \hline
  \small{$\alpha$} & \small{2}  & \small{11} & \small{16}  & \small{21}    \\
    \hline
  \small{$D_{\alpha}(p||q)$} & \small{0.98}  & \small{1.38} & \small{1.43} & \small{1.46} \\
  \small{$D_{\alpha}(p||r)$} & \small{0.05}  & \small{0.15} & \small{0.17} & \small{0.19}  \\
  \hline
  \small{Acceptance} ($\%$) & \small{19.8}  & \small{15.7} & \small{15.1} & \small{13.9}  \\
  \hline

 \end{tabular}
\end{minipage} \\
\vspace{-0.1cm}
\hspace{1cm}
\begin{minipage}{0.45\linewidth}
\centering
\includegraphics[width=3.5cm]{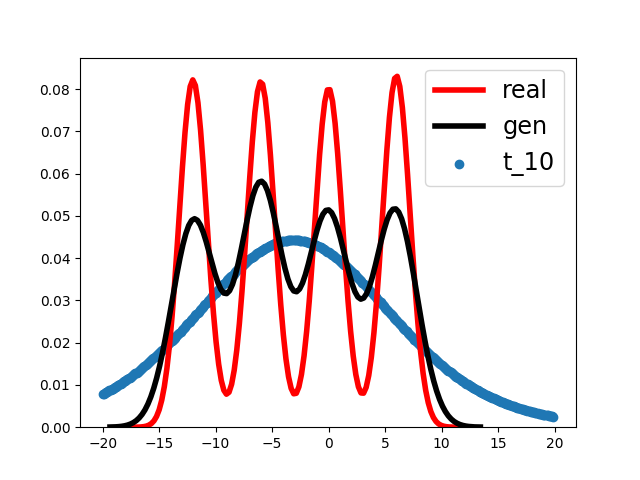}
\label{fig:figure2}
\end{minipage}
\vspace{-12pt}
\caption{\small{Black Plot: Empirical p.d.f. of the generated samples from $\alpha$-DRS algorithm, Red plot: $p(x)$, Blue plot: learned $t$-distribution by RDVI}}
\label{Figure_1}
\vspace{-48pt}
\end{wrapfigure}
\setlength\intextsep{0pt}

The variational distribution $q_{\theta}(x)$ is assumed to be a $t$-distribution with 10 degrees of freedom and parameters $\mu$ and $\log\sigma^{2}$. We have generated 3000 samples from $t$-distribution to approximate $D_{\alpha}(p||q)$. The hyperparameter $T$ was learned using Eq.~\eqref{quantile_funct} ($- \overline{F}(\hat{\theta},\alpha)$) and was used to perform the RS step.




In this case, as evident from Fig.~\eqref{Figure_1}, with the RS step, we are able to get a very good approximation of the target density $p(x)$ despite it having multiple modes. Table ~\eqref{Figure_1} compares the $\alpha$-divergence with RS step ($D_{\alpha}(p||r)$) and without RS step ($D_{\alpha}(p||q)$).



\vspace{15pt}


\subsection{Bayesian Neural Network}
In this section, we will perform approximate inference for Bayesian Neural Network regression. The datasets are collected from the UCI data repository. We have used a single layer NN with 50 hidden units and ReLU activation to model the regression task ~\cite{li2016renyi,wang2018variational}. Let's denote the neural network weights by $\delta$ having a Gaussian prior $\delta\sim \mathcal{N}(\delta;0,I)$. The true posterior distribution of NN weights ($\delta$) is approximated by a fully factorized Gaussian distribution $q(\delta)$.

All the datasets are randomly partitioned 20 times into $90\%$ training and $10\%$ test data. The stochastic gradients are approximated by 100 samples from $q(\delta)$ and a minibatch of size 32 from the training set. We summarize the average RMSE and test log-likelihood in Table ~\eqref{Table_3}. For $\alpha$-DRS method we have chosen acceptance rate to be around 10 $\%$ ($\gamma=0.1$ in equation ~\eqref{quantile_funct}). We have compared the results of $\alpha$-DRS method with RDVI and adaptive f-divergence ~\cite{wang2018variational} ($\beta=-1$).

\vspace{5pt}

\setlength{\tabcolsep}{2pt}

\begin{table} [!htbp]
\begin{tabular}{p{1cm}|c c c| c c c }
\hline
& \multicolumn{3}{c}{R\'enyi $\alpha$ RMSE} & \multicolumn{3}{c}{$\alpha$-DRS RMSE}     \\
\hline
\small{dataset} & $\beta=-1$ & $\alpha=1.0$ & $\alpha =2.0$ & $\beta=-1$ & $\alpha=1.0$ & $\alpha=2.0$  \\
\hline
\small{Boston} & \small{2.861$\pm$ 0.177}     & \small{2.991 $\pm$ 0.198}          &  \small{3.099 $\pm$ 0.196} & \small{\textbf{2.826$\pm$ 0.171}}     & \small{2.900 $\pm$ 0.174}      & \small{2.880 $\pm$ 0.169}   \\
\small{Concrete} & \small{5.343$\pm$ 0.116}   &   \small{5.425 $\pm$ 0.121}         & \small{5.424 $\pm$ 0.105}   & \small{5.292$\pm$0.102}     & \small{\textbf{5.212 $\pm$ 0.110}}       & \small{5.283 $\pm$ 0.111}  \\
\small{Kin8nm} & \small{0.085$\pm$0.001}   & \small{0.084 $\pm$ 0.001}   & \small{0.083 $\pm$ 0.001}    & \small{0.083 $\pm$ 0.001}       & \small{0.082 $\pm$ 0.001}       & \small{\textbf{0.081 $\pm$ 0.001}}   \\
\small{Yacht} & \small{0.810$\pm$0.064}  & \small{1.193 $\pm$ 0.082}            & \small{1.192 $\pm$ 0.089}   &     \small{\textbf{0.772$\pm$0.056}}   & \small{1.082 $\pm$ 0.070} & \small{1.145 $\pm$ 0.081}   \\
\hline

& \multicolumn{3}{c}{R\'enyi $\alpha$ average LL} & \multicolumn{3}{c}{$\alpha$-DRS average LL}     \\
\hline
\small{dataset} & $\beta=-1$   & $\alpha=1.0$ & $\alpha =2.0$ & $\beta=-1$ & $\alpha=1.0$ & $\alpha=2.0$ \\
\hline
\small{Boston} & \small{-2.482$\pm$ 0.177}   &  \small{-2.516 $\pm$ 0.198}        &  \small{-2.549 $\pm$ 0.198}   & \small{\textbf{-2.444$\pm$ 0.171}}    & \small{-2.525 $\pm$ 0.174}      & \small{-2.518 $\pm$ 0.169}  \\
\small{Concrete} & \small{-3.094$\pm$ 0.116}  & \small{-3.107 $\pm$ 0.121}         & \small{-3.10 $\pm$ 0.105}   & \small{-3.082$\pm$ 0.102}      & \small{\textbf{-3.070 $\pm$ 0.110}}      & \small{-3.087 $\pm$ 0.111}  \\
\small{Kin8nm} & \small{1.058 $\pm$ 0.001}     & \small{1.072 $\pm$ 0.001}  & \small{1.084 $\pm$ 0.001}    & \small{1.071$\pm$ 0.001}     & \small{1.093 $\pm$ 0.001}    & \small{\textbf{1.098 $\pm$ 0.001}}    \\
\small{Yacht} & \small{-1.720$\pm$ 0.064}   & \small{-1.959 $\pm$ 0.082}        & \small{-1.977 $\pm$ 0.089}   & \small{\textbf{-1.643$\pm$ 0.056}}      & \small{-1.919 $\pm$ 0.070} & \small{-1.948 $\pm$ 0.081}  \\ 
\hline

 \end{tabular}
 \vspace{0.2cm}
\caption{Test RMSE and Test LL }
\label{Table_3}
\end{table}


\vspace{-1em}
\section{Conclusion}
\label{sec:concl}
\vspace{-1em}
We have presented a two-stage approximate inference method to generate samples from a target distribution. Our approach, essentially a hybrid of R\'enyi divergence variational inference~\cite{li2016renyi} and rejection sampling, leverages a new connection between R\'enyi $\alpha$-divergences and the parameter $M$ controlling the acceptance probabilities of the rejection sampler. Therefore our method can be seen as a rejection sampling-based algorithm that can finetune the variational approximation produced by RDVI into a more expressive sample-based estimate. Our experimental results demonstrate the clear benefits of these improvements in the context of improving variational approximations via rejection sampling.

\bibliography{main}
\bibliographystyle{chicago}

\newpage

\section{Supplementary Material}

In this section, we will show that the approximate Rejection sampling step can further reduce the $\alpha$-divergence between an exact distribution and approximate posterior distribution.

\textbf{Notations}:
\begin{itemize}
    \item True distribution $p(x)=\frac{\tilde{p}(x)}{Z_{p}}$, where $Z_{p}$ is the normalization constant.
    \item Let's denote the learned distribution from $\alpha$-DRS by $r_{\theta}(x)$. We can write this learned distribution as follows:
    \begin{equation}
        r(x)=\frac{q_{\theta}(x)a_{\theta}(x|T)}{Z_{R}(x,T)},\label{resampled_post}
    \end{equation}
    where $Z_{R}(x,T)$ is a normalization constant. For the sake of clarity we will denote $r(x)=\frac{\tilde{r}(x)}{Z_{R}}$, where $Z_{R}$ is a normalization constant.
\end{itemize}

We are making the following assumptions:
\begin{itemize}
    \item The acceptance probability for every sample can be denoted by $a_{\theta}(x|T)$, where $T=-\log M$, $M$ is the constant used for approximate rejection sampling. $T$ can be learned through equation~\eqref{quantile_funct}.
    \begin{eqnarray}
        a_{\theta}(x|T)&=&\text{min}\left[1,\frac{\tilde{p}(x)}{e^{-T}q_{\theta}(x)}\right] \\
        &\approx&\frac{1}{\left[1^{t}+ \left( \frac{e^{-T}q_{\theta}(x)}{\tilde{p}(x)}\right)^{t} \right]^{1/t}}
    \end{eqnarray}
    
\item Take t=1 for getting a differentiable approximation of the acceptance probability.

\end{itemize}

\textbf{Theorem 2}\emph{: For a fixed $\theta$, the approximate Rejection sampling always improves the R\'enyi $\alpha$ divergence between the estimated and actual posterior for $\alpha \in (0,\infty)$. The following equation approximates the acceptance probability.}
\begin{equation}
a_{\hat{\theta}}(x|T)= 1/\left[1 +\left(\frac{q_{\hat{\theta}}(x)e^{-T}}{\tilde{p}(x)}\right) \right], \label{accep_apprx}
\end{equation}

\begin{equation}
    D_{\alpha}(p||r)\leq D_{\alpha} (p||q)  \label{final_deriv1} 
\end{equation}

\begin{itemize}
    \item $T$ $\to$ $\infty$ \emph{implies} $r_{\theta}(x)$ $\to$ $q_{\theta}(x)$
    \item $T$ $\to$ $-\infty$ \emph{implies} $r_{\theta}(x)$ $\to$ $p(x)$
\end{itemize}

\textbf{Proof}:
 We are using the above notations.
\begin{eqnarray}
    D_{\alpha}(P||R)&=&\frac{1}{(\alpha-1)}\log \left[ \int_{}\left(\frac{\tilde{p}(x)}{r(x)}\right)^{\alpha} r(x)dx\right]-\frac{\alpha}{(\alpha-1)}\log Z_{p} \\
          &=& \frac{1}{(\alpha-1)} \left(\alpha\log Z_{R} + \log \left[ \int_{}\left(\frac{\tilde{p}(x)}{\tilde{r}(x)}\right)^{\alpha} r(x)dx\right] \right) -\frac{\alpha}{(\alpha-1)}\log Z_{p} \\
          &=& \frac{\alpha}{(\alpha-1)}\log Z_{R} + \frac{1}{\alpha-1}\log \left[ \int_{}\left(\frac{\tilde{p}(x)}{\tilde{r}(x)}\right)^{\alpha} r(x)dx\right] -\frac{\alpha}{(\alpha-1)}\log Z_{p} 
\end{eqnarray}

Now we will take the derivative of $D_{\alpha}(P||R)$ with respect to T such that variable $T=-\log M$.
\begin{eqnarray}
\nabla_{T} D_{\alpha}(P||R)&=&\frac{\alpha}{(\alpha-1)}\nabla_{T}\log Z_{R}+\frac{1}{\alpha-1} \nabla_{T}\log \left[ \int_{}\left(\frac{\tilde{p}(x)}{\tilde{r}(x)}\right)^{\alpha} r(x)dx\right]\\
&=& \frac{\alpha}{(\alpha-1)}\nabla_{T}\log Z_{R}+\frac{1}{\alpha-1} \frac{\nabla_{T} \int_{}\left(\frac{\tilde{p}(x)}{\tilde{r}(x)}\right)^{\alpha} r(x)dx}{\int_{}\left(\frac{\tilde{p}(x)}{\tilde{r}(x)}\right)^{\alpha} r(x)dx}
\end{eqnarray}

We will take the derivative of numerator separately now for more clarity. Let's denote the numerator by $D_{1}$. Note that the $Z_{R}$ term would be canceled out.

\begin{eqnarray}
    D_{1}&=&\nabla_{T} \int_{}\left(\frac{\tilde{p}(x)}{\tilde{r}(x)}\right)^{\alpha} r(x)dx \\
        &=& -\alpha \int_{}\left(\frac{\tilde{p}(x)}{\tilde{r}(x)}\right)^{\alpha} \nabla_{T}\log \tilde{r}(x) r(x)dx +  \int_{} \left(\frac{\tilde{p}(x)}{\tilde{r}(x)}\right)^{\alpha} \nabla_{T} \log r(x) r(x)dx \\
        &=& -\alpha \nabla_{T} \log Z_{R} \int_{}\left(\frac{\tilde{p}(x)}{\tilde{r}(x)}\right)^{\alpha} r(x) dx +(1-\alpha) \int_{} \left(\frac{\tilde{p}(x)}{\tilde{r}(x)}\right)^{\alpha} \nabla_{T} \log r(x) r(x) dx 
\end{eqnarray}
    
By substituting the above result, we will finally get the following equation.

\begin{equation}
    \nabla_{T}D_{\alpha}(P||R)=-\frac{\int_{} \left(\frac{\tilde{p}(x)}{\tilde{r}(x)}\right)^{\alpha} \nabla_{T} \log r(x) r(x)dx}{\int_{}\left(\frac{\tilde{p}(x)}{\tilde{r}(x)}\right)^{\alpha} r(x)dx} \label{final_deriv} 
\end{equation}

 Since we know that $E_{R}[\nabla_{T}\log r(x)]=0$ we can directly change the numerator above into a covariance function. Also we know that covariance function is unaffected by adding a constant, hence we will add $\nabla \log Z_{R}$ to $\nabla_{T}\log r(x)$ in order to convert it into $\nabla_{T}\log \tilde{r}(x)$. The final derivative would come out to be:

\begin{eqnarray}
 \nabla_{T}D_{\alpha}(P||R)&=&-\frac{\mathrm{COV_{R}}\left[\left(\frac{\tilde{p}(x)}{\tilde{r}(x)}\right)^{\alpha}, \nabla_{T} \log \tilde{r}(x)\right]}{\int_{}\left(\frac{\tilde{p}(x)}{\tilde{r}(x)}\right)^{\alpha} r(x)dx}  \\
 &=& \frac{\mathrm{COV_{R}}\left[\left(\frac{\tilde{p}(x)}{\tilde{r}(x)}\right)^{\alpha},-\left(e^{-T}\frac{\tilde{r}(x)}{\tilde{p}(x)}\right)\right]}{\int_{}\left(\frac{\tilde{p}(x)}{\tilde{r}(x)}\right)^{\alpha} r(x)dx} \\
 &\geq & 0
\end{eqnarray}

Note that in above equation we are taking covariance of a random variable $\left(\frac{\tilde{p}(x)}{\tilde{r}(x)}\right)^{\alpha}$ with its monotonic transformation ($-\left(e^{-T}\frac{\tilde{r}(x)}{\tilde{p}(x)}\right) ,\alpha>0$) which is always positive. Hence, we can conclude that for any general $T$, $D_{\alpha}(P||R)\leq D_{\alpha}(P||Q)$.

\end{document}